\documentclass[11pt,a4paper]{article}
\usepackage[hyperref]{emnlp-ijcnlp-2019}
\usepackage{times}
\usepackage{latexsym}
\usepackage{amsmath}
\usepackage{amssymb} 
\usepackage{latexsym}
\usepackage{graphicx}
\usepackage{hyperref}
\usepackage{booktabs}
\usepackage{multirow}
\usepackage{url}

\aclfinalcopy

\title{Weakly-Supervised Concept-based Adversarial Learning \\
  for Cross-lingual Word Embeddings}

\author{Haozhou Wang \\
  University of Geneva \\
  \And
  James Henderson \\
  Idiap Research Institute\\
  \And
  Paola Merlo \\
  University of Geneva \\}
  
 \date{}

\begin{document}
\maketitle

\begin{abstract}

Distributed representations of words which map each word to a continuous vector have proven useful in capturing important linguistic information not only in a single language but also across different languages. Current unsupervised adversarial approaches show that it is possible to build a mapping matrix that align two sets of monolingual word embeddings together without high quality parallel data such as a dictionary or a sentence-aligned corpus. However, without post refinement, the performance of these methods' preliminary mapping is not good, leading to poor performance for typologically distant languages.

In this paper, we propose a weakly-supervised adversarial training method to overcome this limitation, based on the intuition that mapping across languages is better done at the concept level than at the word level. We propose a concept-based adversarial training method which for most languages improves the performance of previous unsupervised adversarial methods, especially for typologically distant language pairs.

\end{abstract}

\section{Introduction}

Distributed representations of words which map each word to a continuous vector have proven useful in capturing important linguistic information. Vectors of words that are semantically or syntactically similar have been shown to be close to each other in the same space \cite{Mikolov:2013wc,Mikolov:2013uz,Pennington:2014uw}, making them widely useful in many natural language processing tasks such as machine translation and parsing \cite{Bansal:2014ut,Mi:2016vx}, both in a single languages but  also across different  languages.

\citet{Mikolov:2013tp} first observed that the geometric positions of similar words in different languages were related by a linear relation. \citet{Zou:2013ur} showed that a cross-lingually shared word embedding space is more useful than a monolingual space in an end-to-end machine translation task. However, traditional methods for mapping two monolingual word embeddings require high quality aligned sentences or dictionaries \cite{Faruqui:2014vpa,Ammar:2016vt}.

Reducing the need for parallel data, then, has become the main issue for cross-lingual word embedding mapping. Some recent work aiming at reducing resources has shown competitive cross-lingual mappings across similar languages, using a pseudo-dictionary, such as identical character strings between two languages \cite{Smith:2017vw}, or a simple list of numerals \cite{Artetxe:2017wl}.

In a more general method, \citet{Zhang:2017wq} have shown that learning mappings across languages via adversarial training \cite{Goodfellow:2014wp} can avoid using  bilingual evidence. This generality comes at the expense of performance. To overcome this limitation, \citet{Lample:2018wg} refine the preliminary mapping matrix trained by generative  adversarial networks (GANs) and obtain a model that is again comparable to supervised  models for several language pairs. Despite these big improvements, the performance of these refined GAN models depends largely on the quality of the preliminary mappings. It is probably for this reason that these models still do not show satisfactory performance  for typologically distant languages.

We observe that some cross-lingual resources, such as document-aligned data, are not as expensive as high-quality dictionaries. For example, Wikipedia provides thousands of aligned articles in most languages. Based on this observation, in this paper, we explore new augmented models of the adversarial framework that rely on these readily available cross-lingual resources and we develop a weakly supervised adversarial method for learning cross-lingual word embedding mappings.

The main novelty of our method lies in the use of Wikipedia concept-aligned article pairs in the discriminator, which encourages the generator to align words from different languages which are used to describe the same concept. This technique is based on the intuition that the right level of lexical alignment between languages is not the word but the concept. Concepts are then defined as the distribution of words in their respective Wikipedia article, and these distributions are aligned cross-linguistically by the adversarial framework.

The results of our experiments on bilingual lexicon induction show that the preliminary mappings (without post-refinement) trained by our proposed multi-discriminator model are much better than unsupervised adversarial training methods. When we include the post-refinement of \citet{Lample:2018wg}, in most cases, our models are comparable or even better than our dictionary-based supervised baselines, and considerably improve the performance of the method of \citet{Lample:2018wg}, especially for distant language pairs, such as Chinese-English and Finnish-English.\footnote{We provide our definition of {\em similar} and {\em distant} in section 5.} Compared to previous state-of-the-art, our method still shows competitive performance.

\section{Models}
\label{section_models}

The basic idea of mapping two sets of pre-trained monolingual word embeddings together was first proposed by \citet{Mikolov:2013tp}. They use a small dictionary of $n$ pairs of words $\{(w_{s_{_1}}, w_{t_{_1}}), ..., (w_{s_{_n}}, w_{t_{_n}})\}$ obtained from Google Translate to learn a transformation matrix $W$ that projects the embeddings $v_{s_{_i}}$ of the source language words $w_{s_{_i}}$ onto the embeddings $v_{t_{_i}}$ of their translation words $w_{t_{_i}}$ in the target language:

\vspace{-0.5cm}
\begin{equation}
\label{equation_mapping_mikolov}
_{W}^{min}\sum_{i=1}^{n}\| Wv_{s_{_i}} - v_{t_{_i}} \|^{2}
\end{equation}

The trained matrix $W$ can then be used for detecting the translation for any source language word $w_s$ by simply searching a word $w_t$ whose embedding $v_t$ is nearest to $Wv_s$. Recent work by \citet{Smith:2017vw} has shown that enforcing the matrix $W$ to be orthogonal can effectively improve the results of mapping.

\subsection{GANs for cross-lingual word embeddings}
\label{subsection_sgan}

Since the core of this linear mapping is derived from a dictionary, the quality and size of the dictionary can considerably affect the result. Recent attempts by \citet{Zhang:2017wq} and \citet{Lample:2018wg} have shown that, even without dictionary or any other cross-lingual resource, training the transformation matrix $W$ is still possible using the GAN framework \cite{Goodfellow:2014wp}. The standard GAN framework plays a min-max game between two models: a generative model $G$ and a discriminative model $D$. The generator $G$ learns from the source data distribution and tries to generate new samples that appear drawn from the distribution of target data. The discriminator $D$ discriminates the generated samples from the target data.

If we adapt the standard GAN framework to the goal of mapping cross-lingual word embeddings, the objective of the generator $G$ is to learn the transformation matrix $W$ that maps the source language embeddings to the target language embedding space, and the discriminator $D_l$, usually a neural network classifier, detects whether the input is from the target language, giving us the objective:

\vspace{-0.2cm}
\begin{equation}
\label{equation_gan}
\begin{split}
&\min_G \max_D \mathbb{E}_{v_{_t} \sim p_{v_{t}}}logD_l(v_{_t})\\
&+ \mathbb{E}_{v_{_s} \sim p_{v_{_s}}} log(1-D_l(Wv_s)) 
\end{split}
\end{equation}

$D_l(v_{_t})$ denote the probability that $v_t$ came from $p_{v_{t}}$ rather than $p_{v_{s}}$. The inputs of generator $G$ are the embeddings sampled from the distribution of source language word embeddings, $p_{v_{_s}}$, and the inputs of discriminator $D_l$ are sampled from both the real target language word embeddings $p_{v_{_t}}$ and the generated target language word embeddings $p_{Wv_{_s}}$. Both $G$ and $D_l$ are trained simultaneously by stochastic gradient descent. A good transformation matrix $W$ can make $p_{Wv_{_s}}$ similar to $p_{v_{_t}}$, so that $D_l$ can no longer distinguish between them. However, this kind of similarity is at the distribution level. Good monolingual word embeddings normally have a large distribution, and without any post-processing the mappings are usually very rough when we look at specific words \cite{Lample:2018wg}.

\subsection{Concept-based GANs for cross-lingual word embeddings}
\label{subsection_multi_dicrimonator_model}

Under the assumption that words in different languages are similar when they are used to describe the same topic or concept \cite{Sogaard:2015vr}, it seems possible to use concept-aligned documents to improve the mapping performance and make the generated embeddings of the source language words closer to the embeddings of their similar words in the target language. Differently from dictionaries and sentence-aligned corpora, cross-lingual concept-aligned documents are much more readily available. For example, Wikipedia contains more than 40-million articles in more than 200 languages\footnote{https://en.wikipedia.org/wiki/List\_of\_Wikipedias}. Therefore, many articles of different languages can be linked together by the same concept.

Given two monolingual word embeddings $V_s^j = \{v_{s_{_1}}, ..., v_{s_{_j}}\}$ and $V_t^k = \{v_{t_{_1}},, ..., v_{t_{_k}}\}$, our work consists in using a set of Wikipedia concept-aligned article pairs $(C_s^h, C_t^h)=\{(c_{s_{_1}}, c_{t_{_1}}) ..., (c_{s_{_h}}, c_{t_{_h}})\}$ to reshape the transformation matrix $W$. $c_{s_{_i}} \in C_s^h$ represents the article of concept $c_i$ in the source language and $c_{t_{_i}} \in C_t^h$ represents the article of the same concept in the target language.

\begin{figure}
\centering
\includegraphics[scale=0.55]{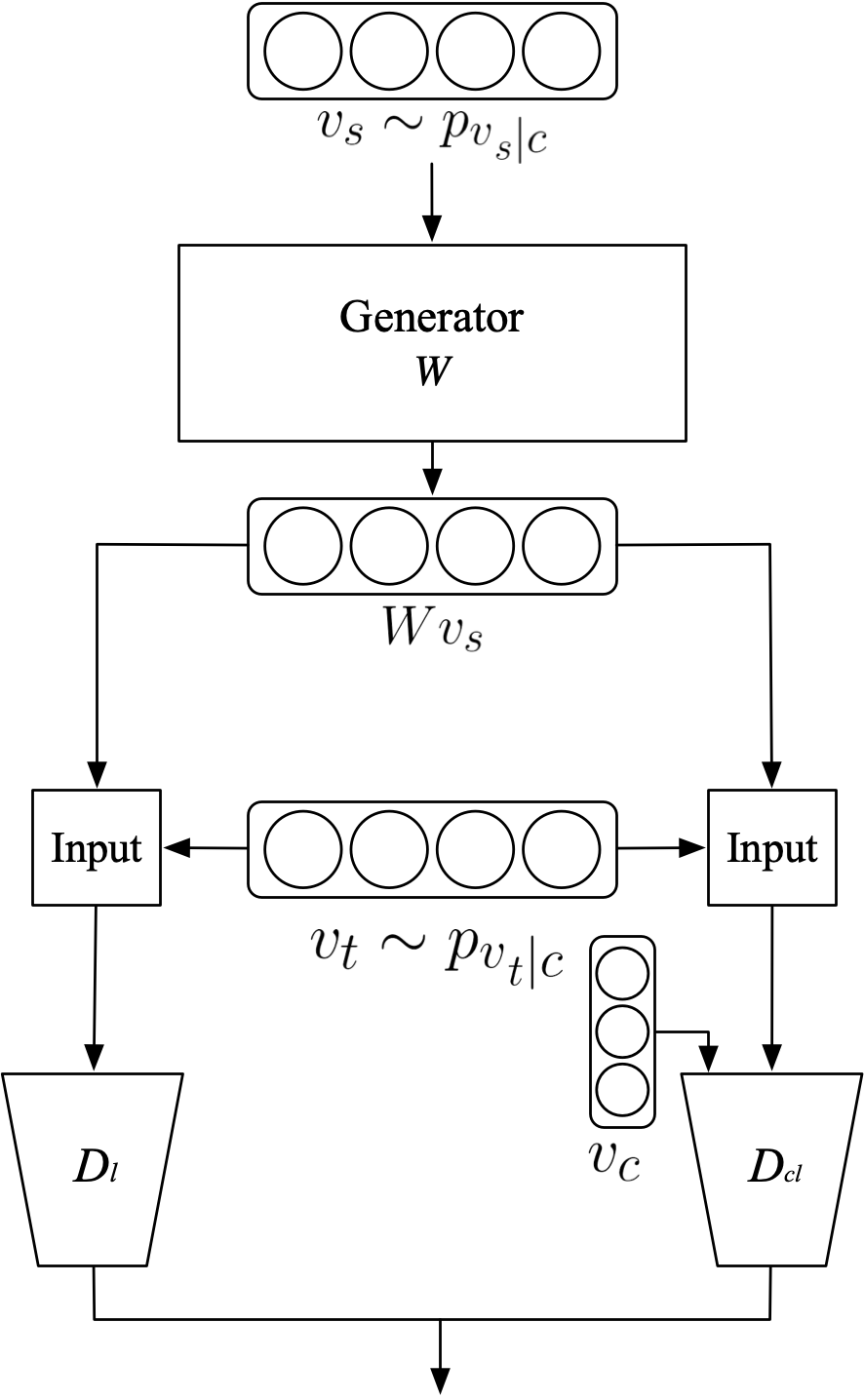}
\vspace{-2ex}
\caption{Architecture of our concept-based multi-discriminator model}
\label{fig_multi_discriminator_model}
\end{figure} 

As shown in Figure~\ref{fig_multi_discriminator_model}, the generator uses the transformation matrix $W$ to map the input embeddings $v_s$ from the source to the target language, differently from the original GANs, the input of the generator and the input of the discriminator on concept $D_{cl}$ are not sampled from the whole distribution of $V_s^j$ or $V_t^k$, but from their sub-distribution conditioned on the concept $c$, denoted $p_v{_{_s|c}}$ and $p_v{_{_t|c}}$. $v_c$ represents the embedding of the shared concept $c$. In this paper, we use the embedding of the title of $c$ in the target language. For titles consisting of multiple words, we average the embeddings of its words.

Instead of determining whether its input follows the whole target language distribution $p_v{_{_t}}$, the objective of the discriminator $D_{cl}$ now becomes to judge whether its input follows the distribution $p_v{_{_t|c}}$ . Because $p_v{_{_t|c}}$ is smaller than $p_v{_{_t}}$ and the proportion of similar words in $p_v{_{_t|c}}$ is higher than in $p_v{_{_t}}$, the embeddings of the input source words have a greater chance of being trained to align with the embeddings of their similar words in the target language.

Our multi-discriminator concept-based model does not completely remove the usual discriminator on language $D_l$ that determines whether its input follows the real target language distribution $p_v{_{_t}}$ or the generated target language distribution, $p_{Wv_{_s}}$. The structure of $D_l$ is very simple, but it is useful when the concept-based discriminator is not stable. The objective function of the multi-discriminator model, shown in equation (\ref{equation_multi_distriminator_model}), combines all these elements.

\vspace{-0.2cm}
\begin{equation}
\label{equation_multi_distriminator_model}
\begin{split}
&\min_G \max_{D_{l}, D_{cl}} \mathbb{E}_{v_{_t} \sim p_{{v_t}|c}} logD_l(v_t)\\
&+ \mathbb{E}_{v_{_s} \sim p_{{v_s}|c}} log(1-D_l(Wv_s))\\
&+ \mathbb{E}_{v_{_t} \sim p_{{v_t}|c}}logD_{cl}(v_t, v_c)\\
&+ \mathbb{E}_{v_{_s} \sim p_{{v_s}|c}}log(1-D_{cl}(Wv_s, v_c))\\
\end{split}
\end{equation}
\vspace{-0.2cm}

\section{Shared Components}
Both the GAN and the concept-based GAN models use a variety of further methods that have been shown to improve results.

\subsection{Orthogonalization}
\label{subsection_orthogonalization}
Previous studies \cite{Xing:2015fn,Smith:2017vw} show that enforcing the mapping matrix $W$ to be orthogonal can improve the performance and make the adversarial training more stable. In this work, we  perform the same update step proposed by \citet{Cisse:2017tm} to approximate setting $W$ to an  orthogonal matrix:

\begin{equation}
\label{equation_orthogonalization}
W\leftarrow \left ( 1+\beta  \right )W -\beta \left ( WW^\top \right )W
\end{equation}

According to \citet{Lample:2018wg} and \citet{Chen:2018wa}, when setting $\beta$ to less than 0.01, the orthogonalization usually performs well.

\subsection{Post-refinement}
\label{subsection_post_refinement}

Previous work has shown that the core cross-lingual mapping can be improved by refining it by bootstrapping from a dictionary extracted from the learned mapping itself \cite{Lample:2018wg}. Since this refinement process is not the focus of this work, we perform the same refinement procedure as \citet{Lample:2018wg} .

After the core mapping matrix is learned, we retrieve the closest target words for the ten thousand most frequent source words. Then, we retrieve back the nearest target words and keep the mutual nearest word pairs for our preliminary dictionary. Finally, we update our preliminary mapping matrix using the objective in equation \ref{equation_mapping_mikolov}. Moreover, following \citet{Xing:2015fn} and \citet{Artetxe:2016ho}, we force the refined matrix $W^*$ to be orthogonal by using singular value decomposition (SVD):

\vspace{-1ex}
\begin{equation}
\label{equation_refine}
W^* = ZU^\top, U\Sigma Z^\top = SVD(V{_s}^\top V{_t})
\end{equation}

\subsection{Cross-Domain  Similarity Local  Scaling }
\label{subsection_csls}

\citet{Radovanovic:2010wb} and  \citet {Dinu:2015jq}  demostrated that
standard nearest  neighbour is not very  effective in high-dimensional
spaces.  Some  vectors  in the target   space  often  appear  as  nearest
neighbours for  many source vectors. Consequently, the target  similar words
retrieved   for a given   source   word  can be   noisy.   The  work   of
\citet{Lample:2018wg} showed that  using cross-domain similarity local
scaling (CSLS)  to retrieve target  similar words is  more accurate
than  standard nearest  neighbours.  Instead of  just considering  the
similarity between  the source word  and its neighbours in  the target
language,  CSLS also  takes into  account the  similarity between  the
target word and its neighbours in the source language:

\vspace{-1ex}
\begin{equation}
\label{equation_csls}
\begin{split}
&CSLS(Wv_s, v_t)=\\
&2cos(Wv_s, v_t)-r_t(Wv_s)-r_s(v_t)
\end{split}
\end{equation}

\noindent
where $r_t(Wv_s)$ represents the mean similarity between a source embedding and its neighbours in the target language, and $r_s(v_t)$ represent the mean similarity between a target embedding and its neighbours in the source language. In this work, we use CSLS to build dictionary for our post refinement.

\begin{table}[]
\renewcommand
\arraystretch{0.8}
\centering
\begin{tabular}{@{}cccc@{}}
\toprule
 & \begin{tabular}[c]{@{}c@{}}Aligned\\ Concepts\end{tabular} & \begin{tabular}[c]{@{}c@{}}Source\\ Vocab.\end{tabular} & \begin{tabular}[c]{@{}c@{}}Target\\ Vocab.\end{tabular} \\ \midrule
de $\rightleftarrows$ en & 0.16M & 1.17M & 0.73M \\
es $\rightleftarrows$ en & 0.15M & 0.61M & 0.65M \\
fi $\rightleftarrows$ en & 0.17M & 0.35M & 0.17M \\
fr $\rightleftarrows$ en & 0.16M & 0.63M & 0.71M \\
it $\rightleftarrows$ en & 0.15M & 0.56M & 0.71M \\
ru $\rightleftarrows$ en & 0.14M & 0.89M & 0.56M \\
tr $\rightleftarrows$ en & 0.01M & 0.24M & 0.19M \\
zh $\rightleftarrows$ en & 0.06M & 0.39M & 0.35M \\ 
\bottomrule
\end{tabular}
\caption{Statistics of Wikipedia concept-aligned articles for our experimentations (M=millions).}
\label{table_data_statistics}
\end{table}

\section{Training}
\label{section_training}

Since our model is trained in a weakly-supervised fashion, the sampling and model selection procedures are important.

\subsection{Sampling Procedure}
\label{subsection_sampling_procedure}

  As the statistics show in Table \ref{table_data_statistics}, even after filtering, the vocabulary of concept-aligned articles is still large. But it has been shown that the embeddings of frequent words are the most informative and useful \cite{Luong:2013wf,Lample:2018wg}, so we only keep the one-hundred thousand most frequent words for learning $W$.
  
For each training step, then, the input word $s$ of our generator is randomly sampled from the vocabulary that is common both to the source monolingual word embedding $S$ and the source Wikipedia concept-aligned articles. After the input source word $s$ is sampled, we sample a concept $c$ according to the frequency of $s$ in each source article of the ensemble of concepts. Then we uniformly sample a target word $t$ from the sub-vocabulary of the target article of concept $c$.\footnote{We use uniform sampling for the target words because the distribution of the sampled sub-vocabulary is very small. Sampling $t$ according to its local frequency in the target article would significantly increase the sampling probability of words like ``the" and ``that", which have less semantic information.}

\subsection{Model Selection}
\label{subsection_model_selection}

\begin{figure}
\centering
\vspace{-1cm}
\includegraphics[scale=0.12]{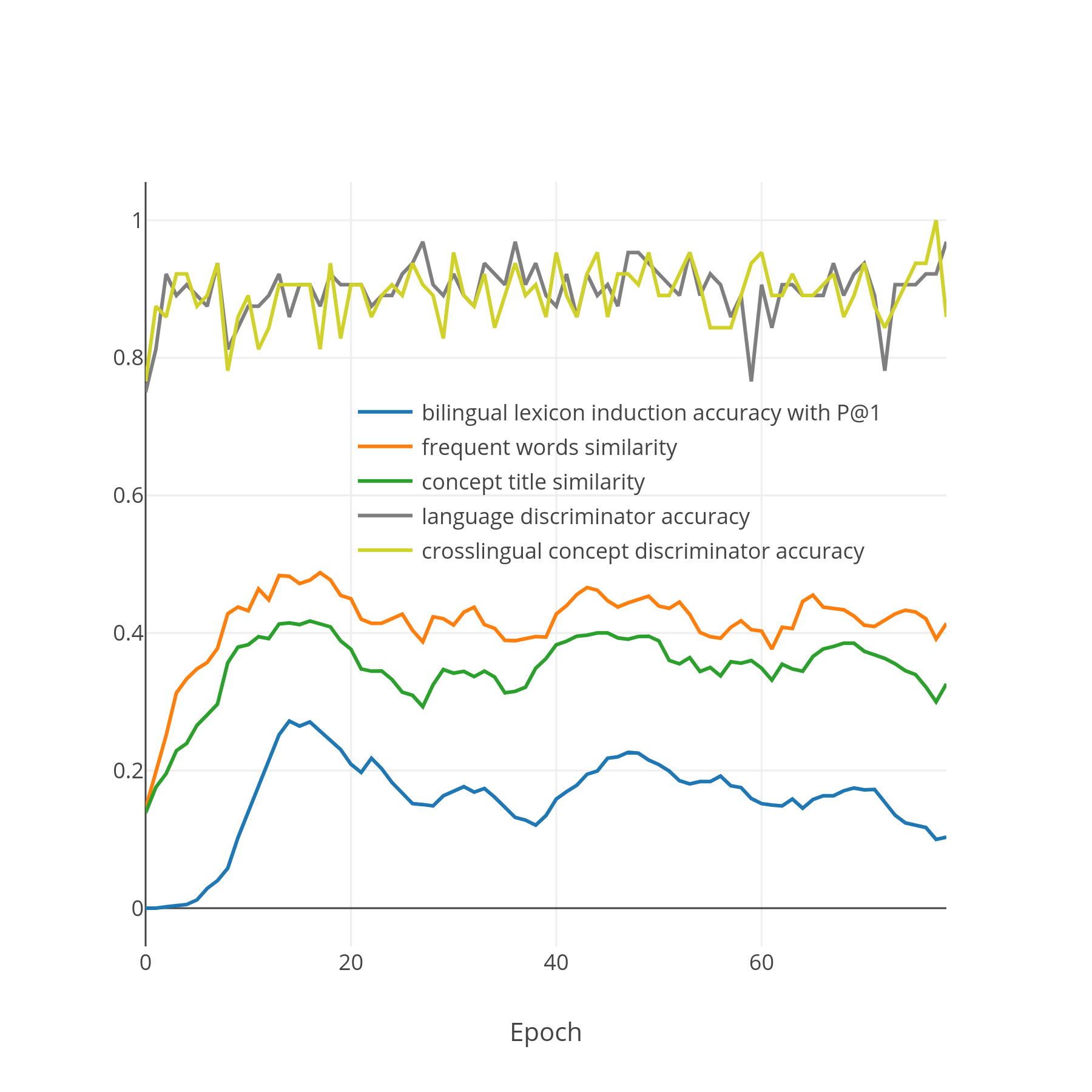}
\vspace{-0.5cm}
\caption{Model selection criteria}
\label{fig_model_selection}
\end{figure}

It is not as difficult to  select the best model for weakly-supervised
learning as  it is for unsupervised  learning. In our task,  the first
criterion  is  the  average  of  the  cosine  similarity  between  the
embeddings  of  the  concept  title  in  the  source  and  the  target
languages. Moreover, we find that  if we record the average embeddings
of the ten most frequent words in the source and target languages for each
aligned article pair, the average  cosine similarity between these two
sets of  embeddings is also a  good indication for selecting  the best
model. As shown in Figure \ref{fig_model_selection}, both of these two
criteria correlate well with the word translation evaluation task that
we describe  in section  \ref{section_experiments}. In this  paper, we
use the average similarity of cross-lingual title as the criterion for
selecting the best model.

\subsection{Other Details}
\label{subscetion_other_details}

For our generator, the mapping matrix $W$ is initialized with a random orthogonal matrix. Our two discriminators $D_l$ and $D_{cl}$ are two multi-layer perceptron classifiers with different hidden layer sizes (1024 for $D_l$ and 2048 for $D_{cl}$), and a ReLU function as the activation function. In this paper, we set the number of hidden layers to 2 for both $D_l$ and $D_{cl}$. In practice, one hidden layer also performs very well; complex classifiers often make the training of models more difficult.

\section{Experiments}
\label{section_experiments}

Since the task of cross-lingual word embedding mapping is situated in
a rich context of related work, in this section, we experimentally
evaluate our proposal on the task of \textbf{Bilingual Lexicon
  Induction} (BLI). This task evaluates directly the bilingual mapping
ability of a cross-lingual word embedding model. For each language
pair, we retrieve the best translations for source words in the test
data, and we report the accuracy.
 
 More precisely, for a given
source language word, we map its embedding onto the target language
and retrieve the closest word. If this closest word is included in the
correct translations in the evaluation dictionary, we consider that it
is a correct case. In this paper, we report the results that use CSLS
to retrieve translations.

Different previous pieces of work on  bilingual lexicon induction use different datasets.  We choose two from the publicly available ones, selected so as to have a comprehensive evaluation of our method.

\paragraph{BLI-1:} The dataset provided by \citet{Lample:2018wg} contains high quality dictionaries for more than 150 language pairs. For each language pair, it has a training dictionary of 5000 words and an evaluation dictionary of 1500 words. This dataset allows us to have a better understanding of the performance of our method on many different language pairs. We choose nine languages for testing and compare our method to our supervised and unsupervised baselines described in section \ref{subsection_baselines}: English(en), German(de), Finish(fi), French(fr), Spanish(es), Italien(it), Russian(ru), Turkish(tr) and Chinese(zh). We classify similar and distant languages based on a combination of structural (directional dependency distance), as proposed and measured in \citet{chen-gerdes2017}  and lexical properties, as measured by the clustering of current large-scale multilingual sentence embeddings.\footnote{See for example the clustering in https://code.fb.com/ai-research/laser-multilingual-sentence-embeddings/}. We consider en $\rightleftarrows$ de, en $\rightleftarrows$ fr, en $\rightleftarrows$ es, en $\rightleftarrows$ it as similar language pairs and en $\rightleftarrows$ fi, en $\rightleftarrows$ ru, en $\rightleftarrows$ tr, en $\rightleftarrows$ zh as distant language pairs.

\paragraph{BLI-2:}
Differently  from BLI-1,  the dataset  of \citet{Dinu:2015jq}  and its
extensions  provided  by \citet{Artetxe:2017wl,  Artetxe:2018vh}  only
consists of  dictionaries of  four language  pairs trained  on a Europarl
parallel corpus.  Each dictionary has  a training set of  5000 entries
and a test set of 1500 entries. Compared to BLI-1, this dataset is much
noisier  and  the  entries   are  selected  from  different  frequency
ranges. However,  BLI-2 has  been widely  used for testing by  previous methods
\cite{Faruqui:2014vpa,Dinu:2015jq,Xing:2015fn,Artetxe:2016ho,Zhang:2016tp,Artetxe:2017wl,Smith:2017vw,Lample:2018wg,Artetxe:2018vx,Artetxe:2018vh}.\footnote{Most
  of these methods have been tested by \citet{Artetxe:2018vh} by using
  their own implementations.}  Using BLI-2 allows us to have a direct
comparison with the state-of-the-art.

\subsection{Baselines:}
\label{subsection_baselines}

As  the objective  of  this  work is  to  improve  the performance  of
unsupervised GANs by using  our weakly-supervised concept-based model,
we choose the  model of \citet{Lample:2018wg} (called  MUSE below), as
our  unsupervised  baseline,  since  it is  a  typical  standard  GANs
model. We evaluate the models both in a setting without refinement and
a setting with refinement. The procedure of refinement is described in
section \ref{subsection_post_refinement} and \ref{subsection_csls}.

Moreover, it is important to  evaluate whether our model is comparable
to previous  supervised models. We use  two different dictionary-based
systems,  VecMap proposed  by  \citet{Artetxe:2017wl} and  Procrustes,
provided     by    \citet{Lample:2018wg},     as    our     supervised
baselines.  Previous experiments  have already  showed that  these two
systems are strong baselines.

\begin{table*}[]
\renewcommand
\arraystretch{0.8}
\centering
\begin{tabular}{@{}cccccccc@{}}
\toprule
\multicolumn{2}{l}{\multirow{2}{*}{}} & \multicolumn{2}{l}{\textit{Supervised}} & \multicolumn{2}{l}{\textit{Without refinement}} & \multicolumn{2}{l}{\textit{With refinement}} \\ \cmidrule(l){3-8} 
\multicolumn{2}{l}{} & VecMap & Procrustes & MUSE & \begin{tabular}[c]{@{}c@{}}Our\\ Method\end{tabular} & Muse & \begin{tabular}[c]{@{}c@{}}Our\\ Method\end{tabular} \\ \midrule
    \multirow{9}{*}{\begin{tabular}[c]{@{}c@{}}Similar\\ Language\\ Pairs\end{tabular}}
 & de-en & \textbf{72.5} & 72.0 & 55.3       & {\em 66.3} & 72.2          & 72.4 \\
 & en-de & 72.5          & 72.1 & 59.2       & {\em 68.2} & 72.5          & \textbf{73.4} \\
 & es-en & 83.4          & 82.9 & {\em 78.1} & 78.0       & 83.1          & \textbf{83.6} \\
 & en-es & \textbf{81.9} & 81.5 & {\em 75.5} & 74.9       & 81.1          & 80.9 \\
 & fr-en & 81.5          & 82.2 & 72.2       & {\em 75.6} & 81.2          & \textbf{83.3} \\
 & en-fr & 81.3          & 81.3 & 77.8       & {\em 78.3} & \textbf{82.2} & \textbf{82.2} \\
 & it-en & \textbf{77.7} & 79.1 & 61.0       & {\em 63.3 }& 76.4          & 76.2 \\
 & en-it & 77.2          & 77.5 & 63.3       & {\em 64.3} & 77.4          & \textbf{78.2} \\\midrule
 & avg. & 78.5           & 78.6 & 67.8       & {\em 71.1} & 78.3          & {\bf 78.8} \\ \midrule
    \multirow{9}{*}{\begin{tabular}[c]{@{}c@{}}Distant\\ Language\\ Pairs\end{tabular}}
 & fi-en & \textbf{55.9} & 56.2          & 0.00 & {\em 42.8} & 28.6 & 53.5 \\
 & en-fi & 39.5          & 39.3          & 0.00 & {\em 35.2} & 21.0 & \textbf{43.0} \\
 & ru-en & 60.5          & \textbf{61.3} & 43.8 & {\em 54.3} & 49.9 & 58.9 \\
 & en-ru & 48.0          & 50.9          & 28.7 & {\em 43.2} & 37.6 & \textbf{51.1} \\
 & tr-en & 55.7          & \textbf{58.0} & 12.2 & {\em 42.1} & 25.5 & 54.7 \\
 & en-tr & 37.3          & 37.1          & 26.2 & {\em 33.1} & 39.4 & \textbf{41.4} \\
 & zh-en & \textbf{45.0} & 41.2          & 26.9 & {\em 32.1} & 30.8 & 42.0 \\
 & en-zh & 45.4          & \textbf{52.1} & 29.4 & {\em 37.7 }& 33.0 & 48.0 \\\midrule
 & avg. & 48.4           & {\bf 49.5}    & 20.9 & {\em 40.1 }& 33.2 & 49.1 \\ \bottomrule
\end{tabular}
\caption{Results of bilingual lexicon induction (accuracy \% P@1) for similar and distant language pairs on the dataset BLI-1. Procrustes AND MUSE represent the supervised and unsupervised model of \citet{Lample:2018wg}, \citet{Artetxe:2017wl} . Word translations are retrieved by using CSLS. Bold face indicates the best result overall and italics indicate the best result between the two columns without refinement.}
\label{table_resualts_lample_data}
\end{table*}

\subsection{Data}
\label{subsection_data}

\paragraph{Monolingual Word Embeddings:} The quality of monolingual word embeddings has a considerable impact on cross-lingual embeddings \cite{Lample:2018wg}. Compared to CBOW and Skip-gram embeddings, FastText \cite{bojanowski2017enriching} capture syntactic information better. 
The ideal situation would be to use FastText embeddings for both the BLI-1 and BLI-2 dataset.  However, much previous work uses CBOW embeddings, so we use
different monolingual word embeddings for BLI-1 and for BLI-2.

For BLI-1, we use FastText\footnote{https://github.com/facebookresearch/fastText} to train our monolingual word embedding models with 300 dimensions for each language and default settings. The training corpus come from a Wikipedia dump \footnote{https://dumps.wikimedia.org/}. For European languages, words are lower-cased and tokenized by the scripts of Moses \cite{Koehn:2007vu}\footnote{http://www.statmt.org/moses/}. For Chinese, we first use OpenCC\footnote{https://github.com/BYVoid/OpenCC} to convert traditional characters to simplified characters and then use Jieba\footnote{https://github.com/fxsjy/jieba} to perform tokenization. For each language, we only keep the words that appear more than five times.

For BLI-2, following the work of \citet{Artetxe:2018vx}\footnote{In the work of \citet{Artetxe:2018vx}\, the authors reimplemented many previous methods described in table \ref{table_resualts_dinu_data}}, we use their pretrained CBOW embeddings of 300 dimensions. For English, Italian and German, the models are trained on the WacKy corpus. The Finnish model is trained from Common Crawl and the Spanish model is trained from WMT News Crawl.

\paragraph{Concept Aligned Data:}

For concept-aligned articles, we use the Linguatools Wikipedia comparable corpus.\footnote{http://linguatools.org/tools/corpora/wikipedia-comparable-corpora/} The statistics of our concept-aligned data are reported in Table \ref{table_data_statistics}.

\subsection{Results and Discussion}
\label{subsection_results_discussion}

Table  \ref{table_resualts_lample_data}  summarizes   the  results  of
bilingual lexicon  induction on  the BLI-1 dataset. We  can see  that with
post-refinement,  our method  achieves the  best performance  on eight
language pairs.  Table \ref{table_resualts_dinu_data}  illustrates the
results of bilingual lexicon induction  on the BLI-2 dataset. Although our
method (with  refinement) does not  achieve the  best results,  the gap
between our method and the best state-of-the-art \cite{Artetxe:2018vx}
is very small,  and in most cases, our method  is better than previous
supervised methods.

\paragraph{Comparison with unsupervised GANs:}

As we have  mentioned before, the preliminary mappings  trained by the
method of \citet{Lample:2018wg} perform well for some similar language
pairs,  such as  Spanish  to  English and  French  to English.   After
refinement, their unsupervised GANs models can reach the same level as
supervised  models  for these  similar  language  pairs. However,  the
biggest  drawback  of   standard  GANs  is  that   they  exhibit  poor
performance  for  distant  language  pairs.  The  results  from  Table
\ref{table_resualts_lample_data}  clearly confirm  this. For  example,
without refinement, the mapping trained by the unsupervised GAN method
can only correctly predict 12\% of  the words from Turkish to English.
Given that  the quality of  preliminary mappings can  seriously affect
the  effect of  refinement, the  low-quality preliminary  mappings for
distant  language pairs  severely limits  the improvements  brought by
post-refinement.    Notice   that    the   method    of
\citet{Lample:2018wg} scores a null result for  English to Finnish on
both BLI-1 and BLI-2, indicating that totally unsupervised
adversarial training can yields rather unpredictable results.

\begin{figure}
\centering
\includegraphics[scale=0.4]{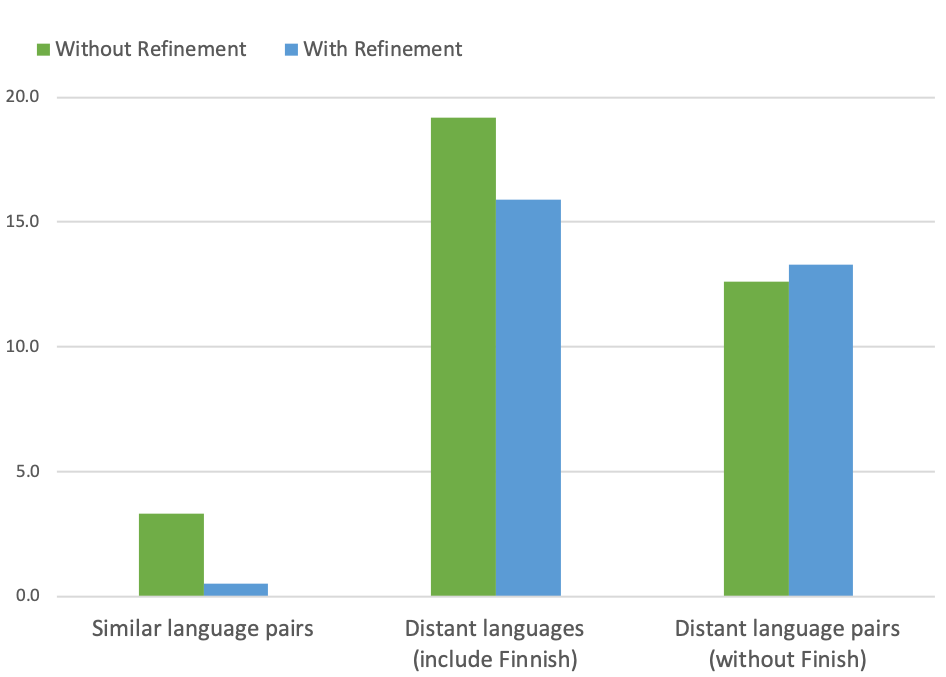}
\vspace{-1ex}
\caption{Average error reduction of our method compared to unsupervised adversarial method for bilingual lexicon induction on BLI-1 dataset \citep{Lample:2018wg}. Since the  Finnish-English pair is an outlier for the unsupervised method, we report both the average with and without this pair.}
\label{figure_average_gap}
\end{figure}

Compared  to the method of  \citet{Lample:2018wg}, the improvement
brought by  our method is apparent.  Our
concept-based GAN  models perform  better  than their  unsupervised
models  on  almost every  language  pairs  for  both BLI-1  and  BLI-2
dataset. For those languages were the better results are not achieved, the
gaps    are   very    small.     
    Figure
\ref{figure_average_gap} reports  the error reduction rate brought
by our method for bilingual lexicon induction on the BLI-1 dataset. It  can be clearly
seen that the improvement brought by  our method is more significant on
distant language pairs than on similar language pairs.

\paragraph{Comparison with supervised baselines:}

Our  two dictionary-based  supervised baselines are  strong. In  most cases,  the
preliminary mappings trained by  our concept-based model are not
comparable with them, but the gab is small.  After post-refinement, our  method becomes comparable  with these
supervised methods. For some language  pairs, such as French-English and
and for English to Finnish, our method performs better.

\paragraph{Comparison with the state-of-the-art:}

\begin{figure}
\centering
\includegraphics[scale=0.13]{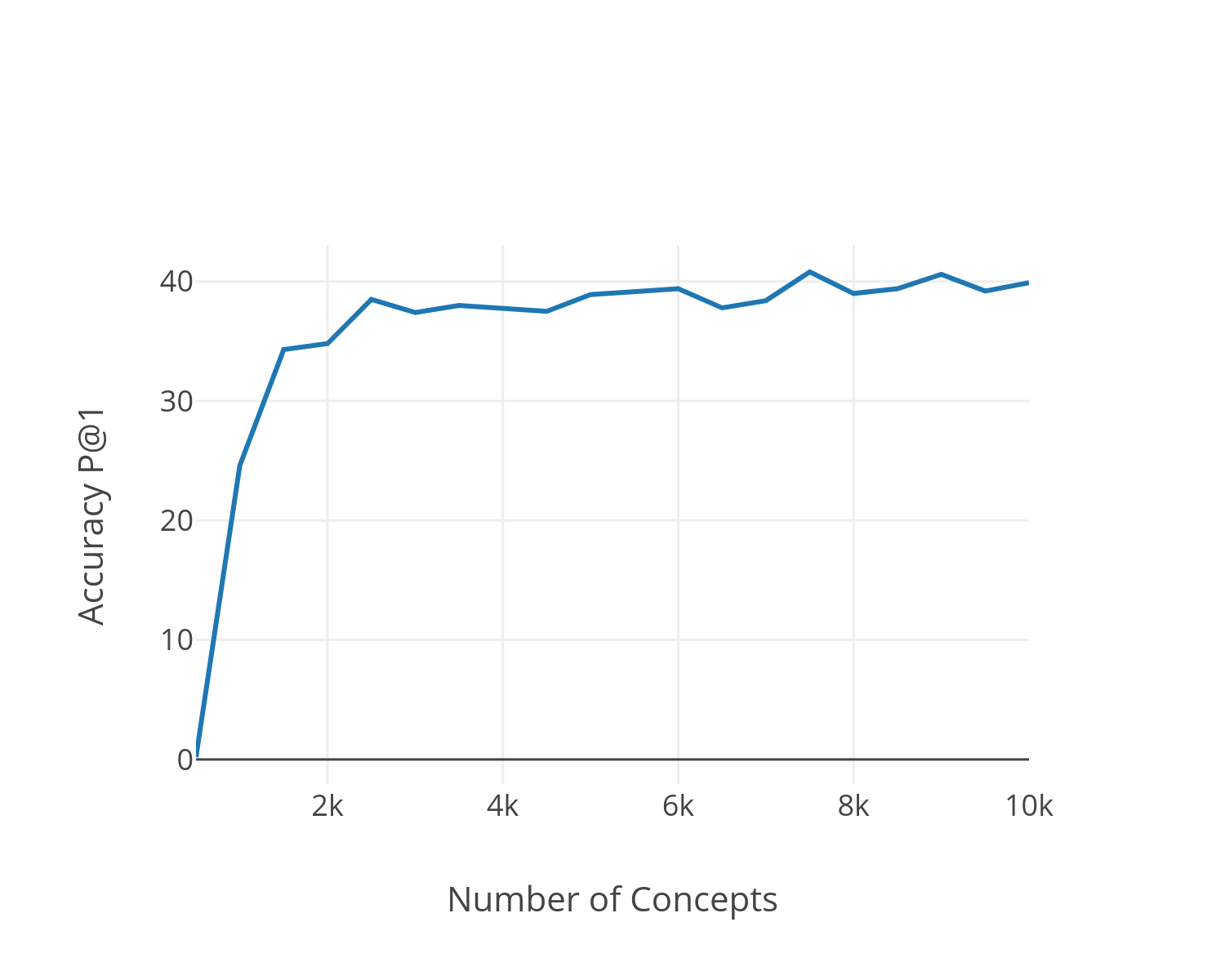}
\vspace{-1ex}
\caption{Accuracy of Chinese-English bilingual lexicon induction task for models trained from different concept numbers.}
\label{figure_concept_number}
\end{figure}

From the results shown in Table \ref{table_resualts_dinu_data}, we can
see  that  in  most  cases,  our method  works  better  than  previous
supervised and  unsupervised approaches.  However, the  performance of
\citet{Artetxe:2018vx} is  very strong  and their method  always works
better than  ours. Two potential  reasons may cause this  difference:
First,  their  self-learning  framework   iteratively  fine-tunes  the
transformation until convergence, while our refinement just runs several
iterations.\footnote{As optimisation of the refinement is not the objective of
  this paper, we  follow the work of \citet{Lample:2018wg} and 
  run  only five iterations  for post-refinement}  Second,  their  framework
consists of many optimization steps,  such as symmetric re-weighting of
vectors, steps that we do not have.
\paragraph{Impact of number of concepts:}
To understand whether the number of aligned concepts affects our method, we trained our concept-based models on a range of Chinese-English concept-aligned article pairs, from 550 to 10'000. We test them on BLI-1 dataset. Following the trend of performance change shown in Figure \ref{figure_concept_number}, we see that when the number of shared concepts reaches 2500, the improvement in accuracy slows down, and is already very close to the result of the model trained from the total number of concepts (Table \ref{table_data_statistics}), thus indicating the direction for further future optimizations.

\begin{table*}[]
\centering
\begin{tabular}{@{}cccccc@{}}
\toprule
 & Method & \textit{en-de} & en-es & \textit{en-fi} & en-it \\ \midrule
\multirow{9}{*}{Supervised} & \citet{Mikolov:2013tp} & 35.0 & 27.3 & 25.9 & 34.9 \\
 & \citet{Faruqui:2014vpa} & 37.1 & 26.8 & 27.6 & 38.4 \\
 & \citet{Dinu:2015jq} & 38.9 & 30.4 & 29.1 & 37.7 \\
 & \citet{Xing:2015fn} & 41.3 & 31.2 & 28.2 & 36.9 \\
 & \citet{Artetxe:2016ho} & 41.9 & 31.4 & 30.6 & 39.3 \\
 & \citet{Zhang:2016tp} & 40.8 & 31.1 & 28.2 & 36.7 \\
 & \citet{Artetxe:2017wl} & 40.9 & - & 28.7 & 39.7 \\
 & \citet{Smith:2017vw} & 43.3 & 35.1 & 29.4 & 43.1 \\
 & \citet{Artetxe:2018vh} & 44.1 & 32.9 & \textbf{44.1} & 45.3 \\ \midrule
\multirow{3}{*}{Unsupervised} & \citet{Zhang:2017wq} & 0.00 & 0.00 & 0.00 & 0.00 \\
 & \citet{Lample:2018wg} & 46.8 & 35.4 & 0.38 & 45.2 \\
 & \citet{Artetxe:2018vx} & \textbf{48.2} & \textbf{37.3} & 32.6 & \textbf{48.1} \\ \midrule
Weakly-supervised & Our method & 47.7 & 37.2 & 30.8 & 46.4 \\ \bottomrule
\end{tabular}
\caption{Results of bilingual lexicon induction (accuracy \% P@1) on BLI-2 dataset, all the results of previous methods come from the paper of \citet{Artetxe:2018vx}}
\label{table_resualts_dinu_data}
\end{table*}

\section{Related Work}
\label{section_related_works}

Sharing a word embedding space across different languages has proven useful for many cross-lingual tasks, such as machine translation \cite{Zou:2013ur} and cross-lingual dependency parsing \cite{Jiang:2015wq,Jiang:2016tz,Ammar:2016ww}. Generally, such spaces can be trained directly from bilingual sentence aligned or document aligned text \cite{Hermann:2014va,ChandarAP:2014wz,Sogaard:2015vr,Vulic:2013vv}. However the performance of directly trained models is limited by their vocabulary size.

Instead of training shared embedding space directly, the work of \citet{Mikolov:2013tp} has shown that we can also combine two monolingual spaces by applying a linear mapping matrix. The matrix is trained by minimizing the sum of squared Euclidean distances between source and target words of a dictionary. This simple approach has been improved upon in several ways: using canonical correlation analysis to map source and target embeddings \cite{Faruqui:2014vpa}; or by forcing the mapping matrix to be orthogonal \cite{Xing:2015fn}.

Recently, efforts have concentrated on  how to limit or avoid reliance
on  dictionaries. Good  results  were achieved  with some  drastically
minimal  techniques.  \citet{Zhang:2016tp}  achieved good  results  at
bilingual POS tagging, but not bilingual lexicon induction, using only
ten word  pairs to build  a coarse orthonormal mapping  between source
and target  monolingual embeddings.  The work  of \citet{Smith:2017vw}
has shown that a singular value decomposition (SVD) method can produce
a  competitive  cross-lingual  mapping by  using  identical  character
strings   across    languages.   \citet{Artetxe:2017wl,Artetxe:2018vh}
proposed  a  self-learning  framework, which  iteratively  trains  its
cross-lingual  mapping  by  using  dictionaries  trained  in  previous
rounds. The initial dictionary of  the self-learning can be reduced to
25  word  pairs  or even  only  a  list  of  numerals and  still  have
competitive  performance.  Furthermore, \citet{Artetxe:2018vx}  extend
their self-learning  framework to  unsupervised models, and  build the
state-of-the-art  for bilingual  lexicon induction.  Instead of  using
a pre-build dictionary for  initialization, they sort the  value of the word
vectors in both  the source and the target distribution, treat two vectors
that have similar permutations as possible translations and
use  them  as  the   initialization  dictionary.  Additionally,  their
unsupervised framework also includes  many optimization augmentations, such
as stochastic dictionary induction, symmetric re-weighting, among others.

Theoretically, employing GANs for training cross-lingual word embedding is also a promising way to avoid the use of bilingual evidence. As far as we know, \citet{Barone:2016go} was the first attempt at this approach, but the performance of their model is not competitive. \citet{Zhang:2017wq} enforce the mapping matrix to be orthogonal during the adversarial training and achieve a good performance on bilingual lexicon induction. The main drawback of their approach is that the vocabularies of their training data are small, and the performance of their models drops significantly when they use large training data. The recent model proposed by \citet{Lample:2018wg} is so far the most successful and becomes competitive with previous supervised approaches through a strong CSLS-based refinement to the core mapping matrix trained by GANs. Even in this case, though, without refinement, the core mappings are not as good as hoped for some distant language pairs. More recently, \citet{Chen:2018wa} extends the work of \citet{Lample:2018wg} from bilingual setting to multi-lingual setting, instead of training crosslingual word embeddings for only one language pair, their approach allows us to train crosslingual word embeddings for many language pairs at the same time. Another recent piece of work which is similar to \citet{Lample:2018wg} comes from \citet{Xu:2018vi}. Their approach can be divided into 2 steps: first, using Wasserstein GAN \cite{Arjovsky:2017vh} to train a preliminary mapping between two monolingual distribution and then minimizing the Sinkhorn Distance across distributions. Although their method performs better than \citet{Lample:2018wg} in several tasks, the improvement mainly comes from the second step, showing that the problem of how to train a better preliminary mapping has not been resolved.

\section{Conclusions}
In this paper, we propose a weakly-supervised adversarial training method for cross-lingual word embedding mapping. Our approach is based on the intuition that mapping across distant languages is better done at the concept level than at the word level. The method improves the performance over previous unsupervised adversarial methods in many cases, especially for distant languages.

\bibliography{m2vec}
\bibliographystyle{acl_natbib}

\end{document}